\documentclass[journal]{IEEEtran}

\usepackage{times}  %Required
\usepackage{helvet}  %Required
\usepackage{courier}  %Required
\usepackage{url}  %Required
\usepackage{graphicx}  %Required
\usepackage{epstopdf}
\usepackage{graphicx}
\usepackage{subfigure}
\usepackage{amsmath,amssymb} % define this before the line numbering.
\usepackage{color}
\usepackage{cite}
\usepackage{booktabs}
\usepackage{threeparttable}
\usepackage{times}
\usepackage{ulem}
\usepackage{xcolor}
\usepackage{subfigure}
\usepackage{slashbox}
\usepackage{multirow}
\usepackage{multicol}
\usepackage{amsmath}
\usepackage{color}
\usepackage{soul}
\usepackage[utf8]{inputenc}
\usepackage[small]{caption}
\usepackage{float}
\usepackage[colorlinks,
            linkcolor=red,
            anchorcolor=blue,
            citecolor=green,
            backref=page
            ]{hyperref}
\usepackage{cleveref}
\begin{document}

\title{Counting from Sky: A Large-scale Dataset for Remote Sensing Object Counting and A Benchmark Method}

\author{Guangshuai Gao,
        Qingjie Liu$^*$,~\IEEEmembership{Member,~IEEE}
        and Yunhong Wang,~\IEEEmembership{Fellow,~IEEE}
\thanks{This work was supported by the National Natural Science Foundation of China (Grant Nos: 41871283, U1804157 and 61601011). (Corresponding author: Qingjie Liu)

Guangshuai Gao, Qingjie Liu and Yunhong Wang are with the State Key Laboratory of Virtual Reality Technology and Systems, Beihang University,
Xueyuan Road, Haidian District, Beijing, 100191, China, and Hangzhou Innovation Institute, Beihang University, Hangzhou, 310051,China (email: gaoguangshuai1990@buaa.edu.cn; qingjie.liu@buaa.edu.cn; yhwang@buaa.edu.cn).} }% <-this % stops a space

\markboth{IEEE Transactions on Geoscience and Remote Sensing}%
{Shell \MakeLowercase{\textit{et al.}}: Bare Demo of IEEEtran.cls for IEEE Journals}

% make the title area
\maketitle

\begin{abstract}
Object counting, whose aim is to estimate the number of objects from a given image, is an important and challenging computation task. Significant efforts have been devoted to addressing this problem and achieved great progress, yet counting the number of ground objects from remote sensing images is barely studied. In this paper, we are interested in counting dense objects from remote sensing images. Compared with object counting in a natural scene, this task is challenging in the following factors: large scale variation, complex cluttered background, and orientation arbitrariness. More importantly, the scarcity of data severely limits the development of research in this field. To address these issues, we first construct a large-scale object counting dataset with remote sensing images, which contains four important geographic objects: buildings, crowded ships in harbors, large-vehicles and small-vehicles in parking lots. We then benchmark the dataset by designing a novel neural network that can generate a density map of an input image. The proposed network consists of three parts namely attention module, scale pyramid module and deformable convolution module to attack the aforementioned challenging factors. Extensive experiments are performed on the proposed dataset and one crowd counting datset, which demonstrate the challenges of the proposed dataset and the superiority and effectiveness of our method compared with state-of-the-art methods.
\end{abstract}


\IEEEpeerreviewmaketitle

\section{Introduction}
\label{sec:intro}

\IEEEPARstart{O}{bject} counting, whose aim is to estimate the number of objects in a still image or video frame, is an important yet challenging computer vision task. It has been attracting increasing research interest because of its practical applications in various fields, such as crowd counting~\cite{zhang2016single,onoro2016towards,boominathan2016crowdnet,kang2018crowd,sam2017switching,sindagi2017generating,liu2018decidenet,hossain2019crowd,zhang2018crowd,sang2019improved,cao2018scale,li2018csrnet,varior2019scale}, cell microscopy~\cite{wang2016fast,walach2016learning,lempitsky2010learning}, counting animals for ecologic studies~\cite{arteta2016counting}, vehicle counting~\cite{zhang2017visual,zhang2017fcn,guerrero2015extremely} and environment survey~\cite{french2015convolutional,zhan2008crowd}. Albeit great progress has been made in many domains of object counting, only a few works have been done to address the problem of ground object counting in remote sensing community over the past few years, for example counting palms or olive trees from remotely sensed images~\cite{bazi2009automatic,salami2019fly,mubin2019young}.

In recent years, with the growing population and rapid development of urbanization, geographic objects such as buildings, cars have become more and more gathering and easily densely crowded. This impels an increasing number of researchers to study the scene understanding from the perspective of object counting. There are emerging studies that analyze crowd scenes from an airborne platform on a helicopter~\cite{bahmanyar2019mrcnet}, and detecting and evaluating the number of cars from drones~\cite{hsieh2017drone} or even satellites~\cite{mundhenk2016large}. However, the dominant ground objects in remote sensing images such as buildings, ships are ignored by the community, estimating the number of which could be beneficial for many practical applications, to name a few urban planning~\cite{rathore2016urban}, environment control and mapping~\cite{pekel2016high}, digital urban model construction~\cite{guan2016digital} and disaster response and assessment~\cite{fan2017quantifying}.

One major reason for this is that there is a lack of large-scale remote sensing object counting dataset available for the community. Although aerial image understanding has gained much attention and there are many datasets built for tasks such as object detection~\cite{xia2018dota,li2020object} or scene classification~\cite{xia2017aid,cheng2017remote}, these datasets are not intended for counting and the number of objects in which is too small to support the counting task. Another reason that limits the progress and applications of counting in remote sensing field lies in that the well-developed counting models may not work well in remote sensing images, since the objects captured overhead possess distinct features from that of captured by surveillance or consumer cameras.

To be specific, compared with the counting task in other fields, e.g., crowd counting, object counting in remote sensing images faces several challenges in the following aspects:

\begin{itemize}
  \item \textbf{Scale variation.} Objects (e.g., buildings) in remote sensing images have large scale variations ranging from only a few pixels to thousands of pixels. This extreme scale change makes it difficult to infer an accurate number of objects.

  \item \textbf{Complex clutter background.} The size of crowded small objects in remote sensing images is so small that it is difficult to detect and recognize them. The objects are often submerged in complex clutter backgrounds, which will distract the models from the region of interest and make them unable to predict the true count.

  \item \textbf{Orientation arbitrariness.} Objects in remote sensing images are taken overhead, thus the orientation is not certain, which is different from the objects in natural images such as crowds have an upright orientation.

  \item \textbf{Dataset scarcity.} As mentioned above, object counting datasets in remote sensing filed are extremely lacking. Although some datasets for object detection in remote sensing images such as SpaceNet\footnote{https://aws.amazon.com/public-datasets/spacenet/} attempt to alleviate this issue, they are not initially designed for the counting task, thus making them not a good choice for the counting research. The recently released datasets such as \mbox{DLR-ACD}~\cite{bahmanyar2019mrcnet} and CARPK~\cite{hsieh2017drone} are acquired using UAV platforms and constructed for crowd or car counting, which is lack of typical geographic objects such as buildings and ships.

\end{itemize}

Intuitively, object counting can be achieved in a straightforward way with the aid of well-designed detectors, such as Faster RCNN~\cite{ren2015faster}, YOLO~\cite{redmon2016you} and SSD~\cite{liu2016ssd}, when the objects are identified and located, the counting itself is trivial. However, this solution could be successful in the condition that objects are with large sizes and sparsely located but may fail in the dense cases, especially for adjoining dense buildings, densely crowded ships in the harbors and small vehicles in the parking lots.

To overcome the aforementioned problems and facilitate future research, we prepare to begin with two routes, one is methodology, and the other is data source. For the methodology, we design a deep supervised network to estimate the count of different geographic objects under complex scenes. Following Lempitsky et al.'s pioneer work~\cite{lempitsky2010learning} that casts the object counting as density estimation, our method also accomplishes the counting task by first predicting the density maps of input images and then integrating them to obtain the final counts. The proposed model is comprised of three stages. The first stage is a truncated VGG-16~\cite{simonyan2014very} serving as a feature extractor that extracts features from the input image, following which is an \textbf{A}ttention module used to highlight informative features and suppress backgrounds. The subsequent stage is a \textbf{S}cale \textbf{P}yramid module with different dilation rates. This module captures the multi-scale information of the objects. The final part is a \textbf{D}eformable convolution layer which we hope the proposed method could be robust to orientation changing. For convenience, we name the proposed method \textbf{ASPD-Net}.

Regarding the data source, we have constructed a large-scale Remote Sensing Object Counting dataset (abbreviated as RSOC). The dataset contains a total of 3057 images and 286,539 instances covered by four widely studied and concerned object types involving buildings, ships, large vehicles, and small vehicles. All instances are accurately annotated with pixel-level center points. To our best knowledge, this is the largest dataset designed for the counting task in remote sensing images. We hope this dataset will motivate researchers of the remote sensing community to pay attention to the counting topic and promote research on aerial scene understanding.

In summary, we make the following contributions.
\begin{enumerate}
  \item We propose a specially designed deep neural network ASPD-Net to attack the challenges of object counting in remote sensing images. The proposed ASPD-Net addresses the problems of scale and rotation variation by introducing a feature pyramid module and deformable convolution module. In addition, the attention mechanism is incorporated to jointly aggregate multi-scale features and suppress clutter backgrounds.

   \item A large-scale remote sensing object counting dataset, termed RSOC, is constructed and released to the community to boost the development of object counting in the field of remote sensing. The RSOC dataset consists of 3,057 images with 286,539 annotations and covers four categories, including buildings, ships, large-vehicles, and small-vehicles. To the best of our knowledge, this is the first attempt to build and release a dataset that facilitates research of object counting in remote sensing images.

  \item We set a benchmark for the RSOC dataset by conducting extensive experiments, which demonstrates the effectiveness of our proposed method. We also make comparisons with several state-of-the-art counting methods, and the proposed method achieves superior performance on the RSOC dataset. Additionally, experiments on one widely used crowd counting dataset demonstrate the robustness and generalization ability of the proposed approach.
\end{enumerate}

This paper extends and makes further improvement compared with its previous version~\cite{gao2020counting} in the following aspects: 1) More details and descriptions of the dataset, including the collection of the data, annotation manner and statistical information are added into this paper; 2) We give a brief review of related work to enable readers to have a comprehensive understanding of our work; 3) More experiments including ablation studies and comparisons with state-of-the-art counting methods are included to demonstrate the effectiveness and superiority of our approach.

The remainder of this paper is organized as follows. In Section~\ref{section:related} we briefly review some works related to this paper. The dataset is presented in  Section~\ref{section:dataset}. We give detailed structures of the proposed method and experiments in Section~\ref{section:method} and Section~\ref{section:experimets}, respectively. Finally, this paper is concluded in Section~\ref{section:conclusion}.

% --------------------------------------------------------------------------------------------------------------------------%
\section{Related work}
\label{section:related}
\subsection{Object counting}
Contemporary works for object counting can be roughly classified into three categories: 1) detection-based; 2) regression-based and 3) density map estimation based methods. Detection-based methods are inchoate counting techniques, which estimate the number of objects from the detection of object instances. The performance of these methods relies on sophisticated detectors and is not satisfactory owing to the fact that object detection is still in its primitive stage in the early years. In recent years, with the advent of deep learning techniques, object detection has achieved significant progress. Object detectors such as Faster RCNN~\cite{ren2015faster}, SSD~\cite{liu2016ssd}, YOLO~\cite{redmon2016you} have shown remarkable performance in various object detections such as face~\cite{yang2016wider}, pedestrian~\cite{dollar2012pedestrian} and so on. Counting is completed spontaneously after the detection by simply summarizing the number of detection instances. However, these approaches can only work well in the easy case, i.e., objects are sparsely located and in relatively large scales, they may show poor performance in densely congested scenes, particularly in remote sensing images, since ground objects such as dense buildings, small-vehicles, large-vehicles, and ships are far beyond current detection performance.

Regression-based methods, also known as global-regression based methods, count objects via mapping the high-dimension images into the real number space through regression models such as linear regression~\cite{paragios2001mrf} or Gaussian mixture regression~\cite{tian2010latent}. Elaborately designed hand-crafted features such as SIFT~\cite{lowe1999object}, GLCM~\cite{haralick1973textural} are usually employed to represent images. These methods are successful when dealing with objects with homogeneity scales and uniform distributions, however, may fail when objects are with varying scales and cluster distributions.

Lempitsky et al.~\cite{lempitsky2010learning} set a milestone for subsequent counting researches by casting the visual object counting problem as image density estimation, in which the integral of the density map gives the count of objects within it. Following this work, Pham et al.~\cite{pham2015count} learn a non-linear mapping function by using random forest regression. Recently, with the entering of the deep learning era, many CNN based counting methods have been proposed and achieved great success, especially in crowd counting. Zhang et al.~\cite{zhang2016single} propose a multi-column neural network (MCNN) with three branches, each of which constructed with different kernel sizes for capturing varisized objects. Sindagi et al.~\cite{sindagi2017cnn} improve the MCNN by jointly learning crowd count classification and density map estimation, and integrate them into an end-to-end cascaded network. Sam et al.~\cite{sam2017switching} propose a switching CNN to select the most suitable regressor by training a switch classifier. Instead of designing a wider multi-column network, Li et al.~\cite{li2018csrnet} propose to devise a deeper single column network that utilizes dilated convolution to enlarge the receptive fields so as to boost counting performance. In this paper, we also intend to design a CNN based model for estimating density maps of remote sensing images by taking advantage of recently developed techniques such as attention mechanism.

\subsection{Attention mechanism}
Attention mechanism has been incorporated into deep neural networks for improving performance in various computer vision tasks, including image captioning~\cite{xu2015show,you2016image}, image question answering~\cite{yang2016stacked}, video analysis~\cite{huang2019attentive}, image classification~\cite{wang2017residual}, object counting~\cite{liu2018decidenet,zhang2019attentional} and countless others. Bahdanau et al.~\cite{bahdanau2014neural} are among the first to introduce the attention mechanism and successfully apply it to machine translation. Chen et al.~\cite{chen2017sca} propose to encode the spatial- and channel-wise attention sequentially for improving image captioning performance. Wang et al.~\cite{wang2018non} put forward a non-local neural network to compute the response at a position as a weighted sum of feature maps. Woo et al.~\cite{woo2018cbam} devise a convolution block attention module (CBAM) to enrich feature representations. It can be plugged into many feed-forward convolutional networks. The attention mechanism has also been introduced into the counting field. For instance, Liu et al.~\cite{liu2018decidenet} incorporate an attention module to adaptively decide the appropriate counting mode for different locations on the image based on its real density conditions. Zhang et al.~\cite{zhang2019attentional} propose an attentional neural field network, in which they introduce the no-local attention module~\cite{wang2018non} to expand the receptive field to the entire image such that the method can deal with large scale variations. Our work arranges the channel and spatial attention modules in different manners. The rational arrangement will be demonstrated in Section~\ref{subsection:arrangement}. With the attention module incorporated in our proposed remote sensing object counting framework, we can achieve the goal of highlighting the region of interest and alleviating the interference of cluttered backgrounds.

\subsection{Dilated convolutions}
Dilated convolution has been widely used in a bunch of vision tasks, due to its prominent characteristics of enlarging the receptive field of networks without increasing the computation complexity. Yu et al.~\cite{yu2015multi} design a dilated convolution-based network to capture multi-scale contextual features for better segmenting objects. Song et al.~\cite{song2018pyramid} leverage pyramid dilated convolution for video salient object detection. Li et al.~\cite{li2018recurrent} utilize dilated convolution for image de-raining. In counting community, CSRNet~\cite{li2018csrnet} also employs dilated convolution to design a deep convolutional neural network for crowd counting in congested scenes.

Different from previous approaches, our proposed work incorporates the attention mechanism to capture more contextual features and then concatenates several parallel dilated convolutions with different dilation rates to extract multi-scale features. Furthermore, a set of deformable convolution layers is leveraged to generate high-quality density maps to accurately locate the position of objects, to address the issue that orientation arbitrariness due to overhead perspective in the remote sensing images.

% ----------------------------------------------------------------------------------------------------------------------------%

\section{Methodology}
\label{section:method}

\begin{figure*}[!htb]
	\centering
	\centerline{\includegraphics[width=1.0\textwidth]{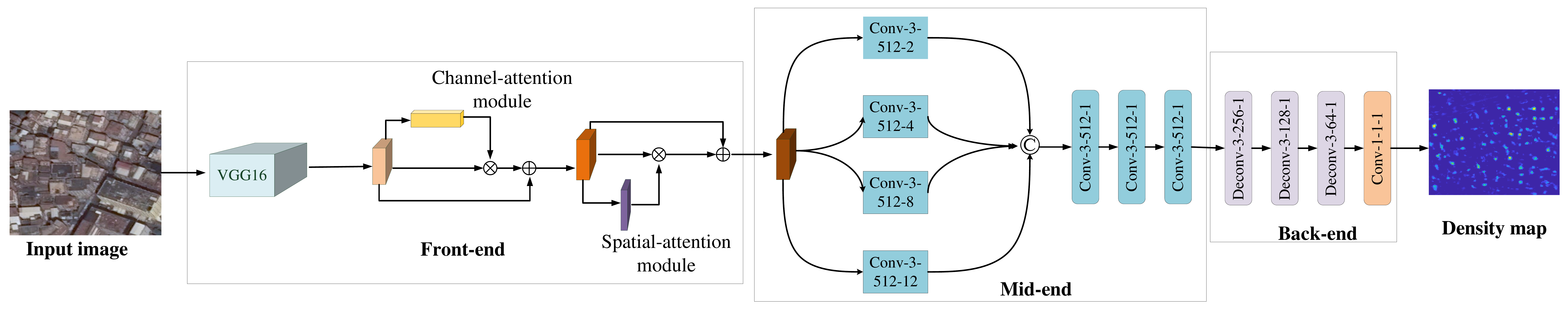}}
	\caption{Architecture of ASPD-Net. The parameters of convolutional layers in the the mid-end stages are denoted as "Conv-(kernel size)-(number of filters)-(dilation rate)", while the parameters in the back-end stage are represented as "Conv-(kernel size)-(number of filters)-(stride)".}
	\label{figure:architecture}
\end{figure*}

The architecture of the proposed ASPD-Net is illustrated in Fig.~\ref{figure:architecture}. It comprises of three stages. The front-end is a truncated VGG-16~\cite{simonyan2014very} incorporated with an attention module to extract features of an input image. The mid-end is a scale pyramid module (SPM) followed by four dilated convolution layers to deal with scale variation. Finally, we equip the back-end with a set of deformable convolution layers for better capturing orientation information and add a $1 \times 1$ convolution layer to generate density map.

\subsection{Feature extraction with attention (front-end)}
For a given remote sensing image with an arbitrary size, we first feed it into a feature extractor, which is composed of the first 10 convolution layers of the pre-trained VGG-16~\cite{simonyan2014very}. The VGG-16 is widely used in counting tasks as backbone to build models for its strong generalization ability~\cite{sam2017switching,sindagi2017generating,li2018csrnet}, thus in this work we also employ VGG-16 as the basis to construct our network. VGGs were initially designed and trained for image classification task~\cite{russakovsky2015imagenet}. Although it has been proved that VGGs could generalize well to remote sensing images after fun-tuning~\cite{liu2018classifying}, additional designs should be considered to address counting tasks, since there is a significant gap between counting and classification. The features should be enhanced to better represent crowded objects.

Inspired by recent development in attention mechanism, especially \cite{woo2018cbam} and \cite{gao2019scar}, we incorporate attention modules to achieve the goal of capturing more contextual and high-level semantics. We consider both channel- and spatial-attention, which are described below:

\begin{figure}[!htb]
	\centering
	\centerline{\includegraphics[width=0.5\textwidth]{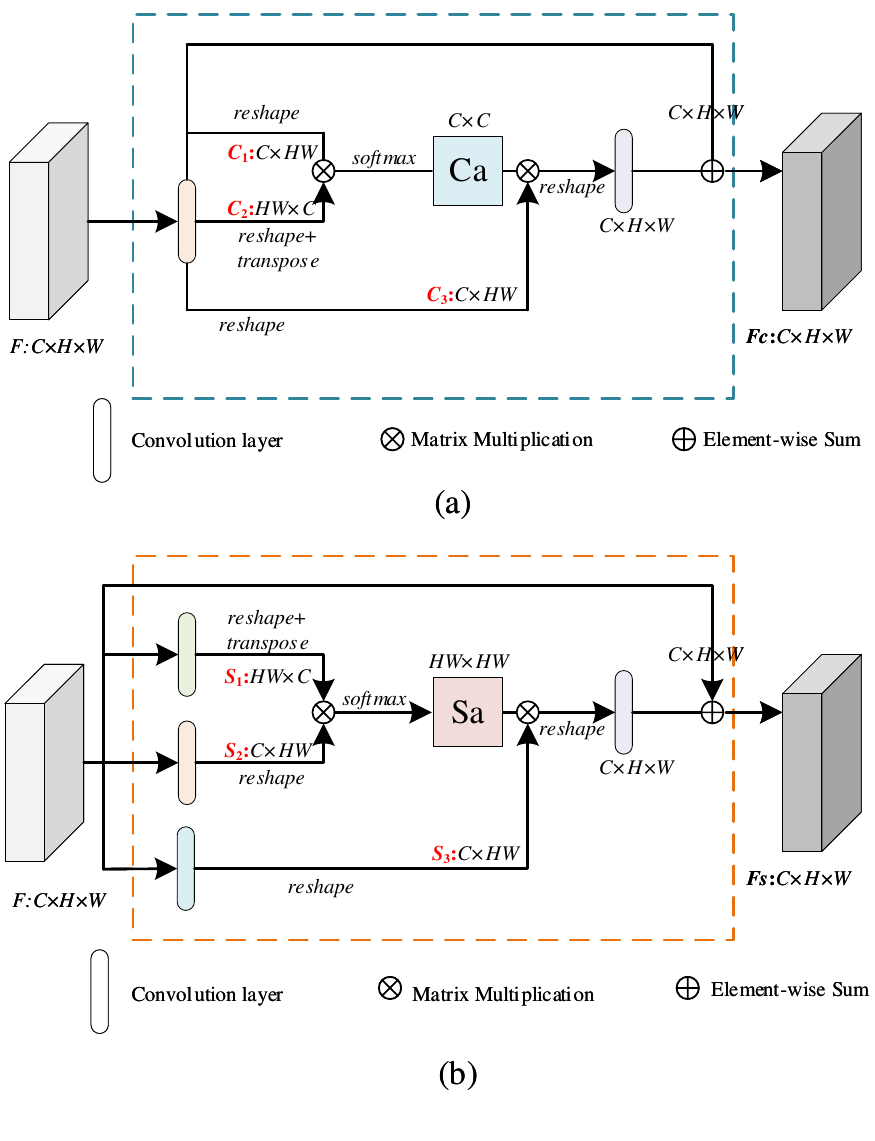}}
	\caption{The detailed architecture of (a) channel-attention and (b) spatial-attention.}
	\label{figure:cbam}
\end{figure}

\subsubsection{\textbf{Channel-attention}}
In the extremely dense scenes, textures of the foregrounds are hard to distinguish from that of the backgrounds. Channel-attention could alleviate this problem. The architecture of channel-attention is depicted in Fig.~\ref{figure:cbam}(a).

Concretely, for a given intermediate feature map $\mathbf{F} \in \mathbb{R}^{C \times H \times W}$, where $C$ represents the number of the channels, $H$ and $W$ denote the height and width of the feature map. First of all, one $1\times1$ convolution layer is executed, and then through reshaping or transpose operations, two feature maps $C_{1}$ and $C_{2}$ are obtained. To generate the channel attention map, a matrix multiplication and softmax operations are applied on $C_{1}$ and $C_{2}$. In this way, we obtain a channel attention map $C_{a}$ with a size of $C \times C$. Specifically, the process can be formulated as follows:

\begin{equation}
C_{a}^{j i}=\frac{\exp \left(C_{1}^{i} \cdot C_{2}^{j}\right)}{\sum_{i=1}^{C} \exp \left(C_{1}^{i} \cdot C_{2}^{j}\right)}
\end{equation}
where $C_{a}^{j i}$ means the influence of $i$th channel on the $j$th channel. The final weighted feature maps with channel attention module with size of $C \times H \times W$ is calculated as:

\begin{equation}
C_{\text {final}}^{j}=\lambda \sum_{i=1}^{C}\left(C_{a}^{j i} \cdot C_{3}^{i}\right)+F^{j}
\label{equ:ca}
\end{equation}
where $\lambda$ is a learnable parameter, which can be leaned from a $1\times1$ convolution operation.
\subsubsection{\textbf{Spatial-attention}}
Observing that there have different densities in different locations of the feature map, we further encode long-range dependencies from spatial dimension, which is effective for the performance of the spatial locations. Similar to the operations in the channel-attention aforementioned, the architecture of the spatial-attention is illustrated in Fig.~\ref{figure:cbam}(b). However, spatial-attention has some differences from channel-attention in two folds:
1) Instead of only one $1\times 1$ convolution layer adopted in channel-attention, it has three in the spatial attention operations; 2) In contrast to the size of the channel attention map ($C_{a}$) is $C \times C$, the size of spatial attention map ($S_{a}$) is $HW \times HW$. Specifically, $S_{a}$ can be computed as follows:
\begin{equation}
S_{a}^{j i}=\frac{\exp \left(S_{1}^{i} \cdot S_{2}^{j}\right)}{\sum_{i=1}^{H W} \exp \left(S_{1}^{i} \cdot S_{2}^{j}\right)}
\end{equation}
where $S_{a}^{ji}$ denotes the influence at the position of the $i$th pixel on the $j$th pixel of the feature map. More similar of the positions means stronger correlations between them. Then the final weighted feature map with spatial attention whose size is $C \times H \times W$ can be obtained as below:
\begin{equation}
S_{\text {final}}^{j}=\mu \sum_{i=1}^{H W}\left(S_{a}^{j i} \cdot S_{3}^{i}\right)+F^{j}
\label{equ:sa}
\end{equation}
where $\mu$ is a learnable parameter, which is leaned with the same operations as $\lambda$ in Equ.~\ref{equ:ca}.
%-----------------------------------------------------------------------------------------------------------------------------------------------------------------------------------------%
\subsection{Scale pyramid module (mid-end)}
There are multiple max-pooling operations in the front-end stage, leading to dramatically decrease in the size of the feature map. The size of the output map is only 1/64 of the input image. This will bring two drawbacks. Firstly, a density map indicates the localization of objects in the image. Pooling operation makes the features invariance to local translations thus is good for classification however is harmful for object localization, in here is a barrier to generating high-quality density map. Secondly, the information of small objects is weakened as the spatial resolution of the feature map decreases, making the network blind to the small objects.

\begin{figure}[!tb]
	\centering
	\centerline{\includegraphics[width=0.5\textwidth]{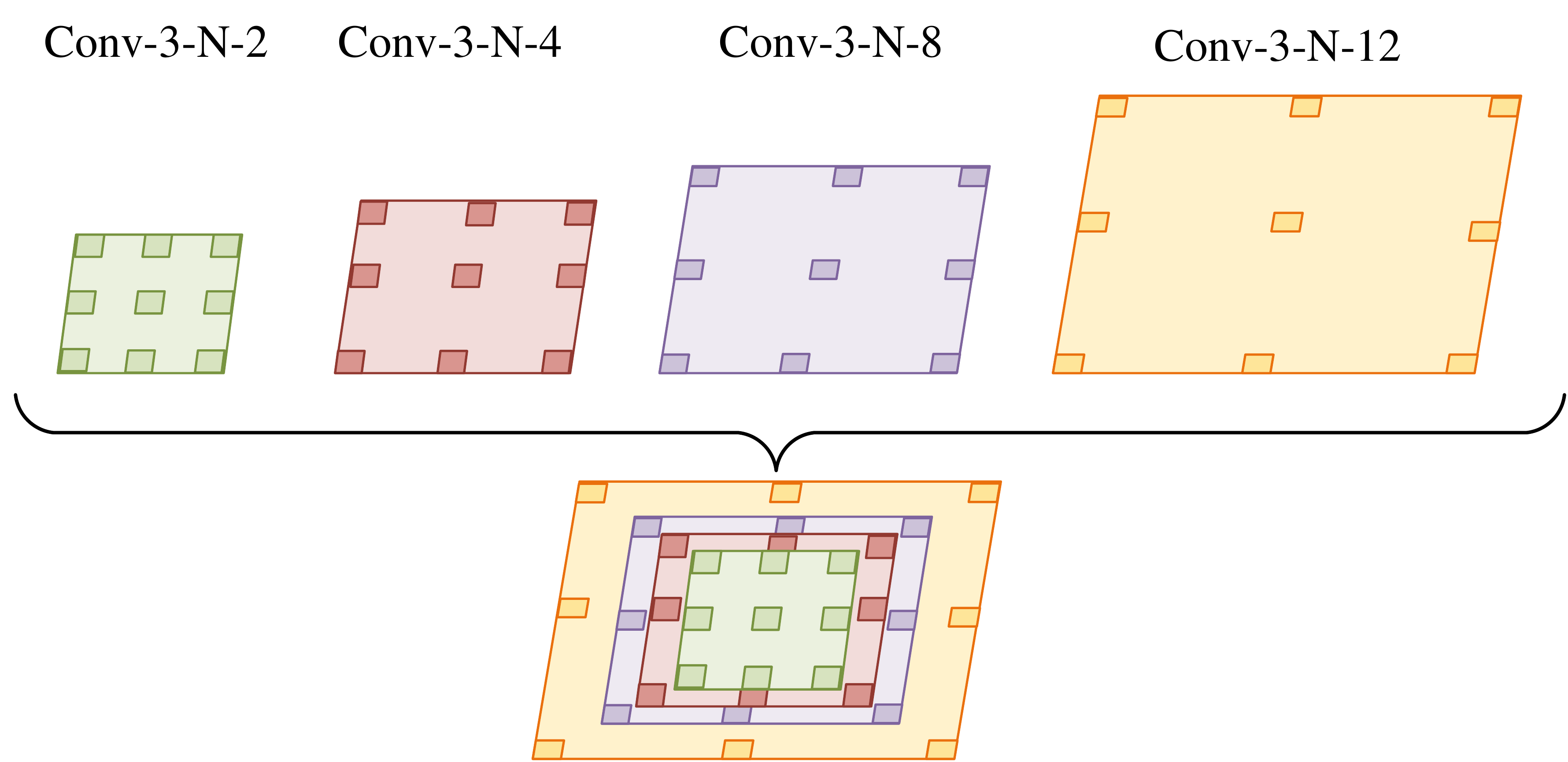}}
	\caption{The structure of SPM. The parameters of convolutional layers are denoted as ``Conv-(kernel size)-(number of filters)-(dilation rate)".}
	\label{figure:spm}
\end{figure}

To address these problems, inspired by \cite{chen2019scale}, we introduce a scale pyramid module (SPM) that is built with four parallel dilated convolution layers. The dilated convolution is convolution with holes as illustrated in Fig.~\ref{figure:spm}. It was first introduced by Yu et al.~\cite{yu2015multi} in the segmentation task to expand the receptive fields without losing spatial resolution of feature maps. In addition, there are no extra parameters or computations, making it an excellent solution for dense prediction tasks, e.g., scene parsing~\cite{zhao2017pyramid} and object counting~\cite{li2018csrnet,chen2019scale}.

In this work, all the four layers have the same channels however different dilation rates to capture different scale information. We use dilation rates 2, 4, 8 and 12 as suggested in \cite{chen2019scale}. In this way, we build a pyramid with different receptive fields that can not only keep the spatial resolution of feature maps unchanged but also can be invariant to scale variations. The structure of the SPM is depicted in Fig.~\ref{figure:spm}.

\subsection{Deformable convolution module (back-end)}
\begin{figure}[!tb]
	\centering
	\centerline{\includegraphics[width=0.5\textwidth]{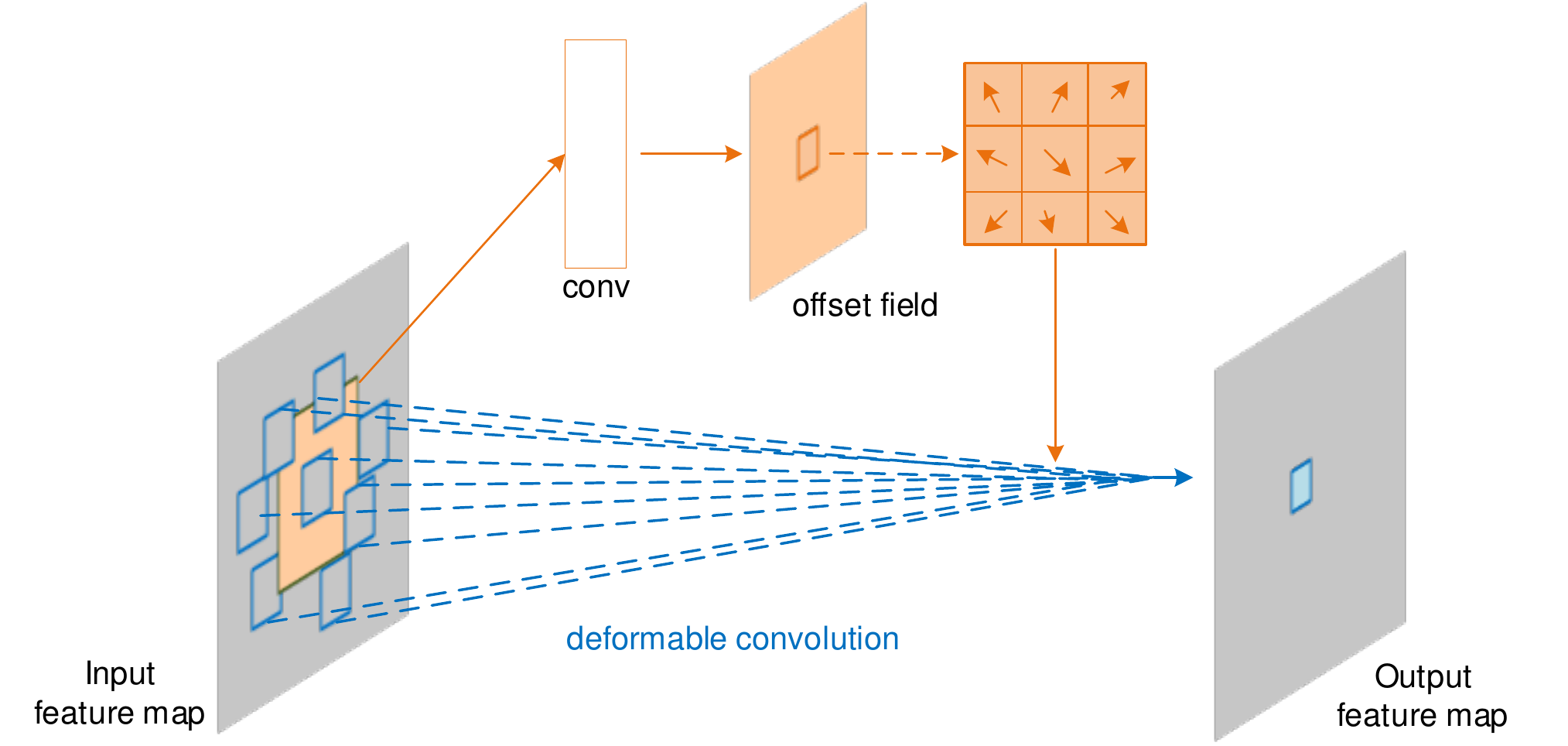}}
	\caption{Diagram of deformable convolution module~\cite{dai2017deformable}.}
	\label{figure:deform1}
\end{figure}

\begin{figure}[!tb]
	\centering
	\centerline{\includegraphics[width=0.45\textwidth]{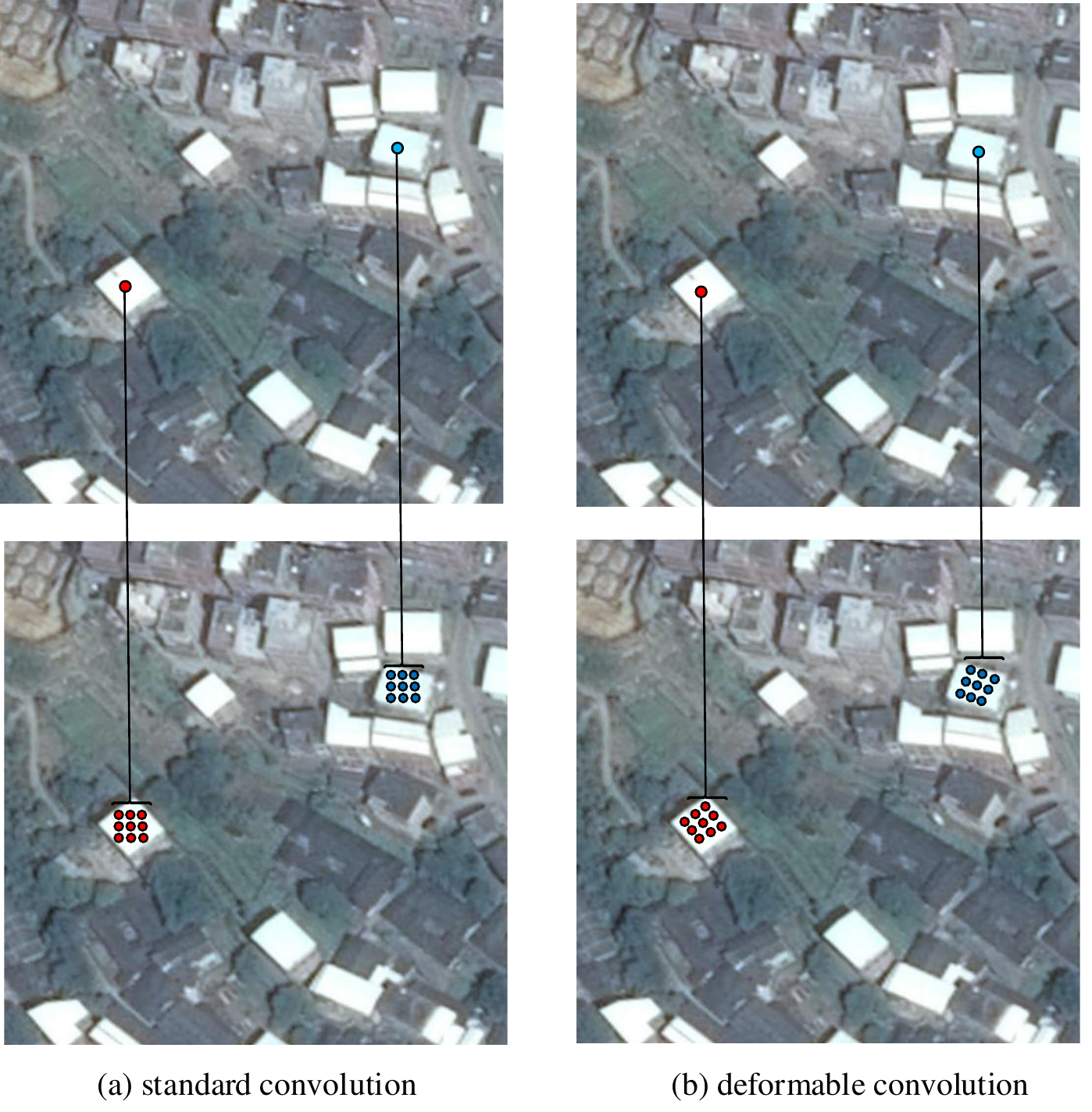}}
	\caption{Visualization of deformable sampling locations. (a) standard convolution (b) deformable convolution.}
	\label{figure:deform2}
\end{figure}

Deformable convolution~\cite{dai2017deformable} is an operation that adds an offset, whose magnitude can be learnable, on each point in the receptive field of the feature map. After the offset, the shape of the receptive field matches the actual shape of the object rather than a square. The advantage of the offset is that no matter how deformable the object is, the region of convolution always covers the object. The diagram and the visualization of deformable convolution are illustrated in Fig.~\ref{figure:deform1} and Fig.~\ref{figure:deform2}, respectively.

For a normal convolution, a given sampling location $\mathbf{p}_{m}$ with $3\times3$ convolution kernel of dilation 1, which can be represented as $\mathbf{p}_{m} \in \mathcal{M}=\{(-1,-1),(0,-1).  \ldots,(1,0),(1,1)\}$. And then the output feature map $\mathbf{y}(\mathbf{p})$ on the location $\mathbf{p}$ can be formulated as

\begin{equation}
\mathbf{y}(\mathbf{p})=\sum_{m=1}^{\mathcal{M}} \mathbf{w}\left(\mathbf{p}_{m}\right) \cdot \mathbf{x}\left(\mathbf{p}+\mathbf{p}_{m}\right)
\end{equation}
where $\mathbf{w}$ represents weighted parameters and $\mathbf{x}$ means the input feature map.

In contrast with normal convolution, deformable convolution adds an adaptive learnable offset $\Delta \mathbf{p}_{m}$, which can be optimized via training~\cite{dai2017deformable}. Therefore, the deformable convolution of feature map $\mathbf{y}(\mathbf{p})$ can be represented as

\begin{equation}
\mathbf{y}(\mathbf{p})=\sum_{m=1}^{\mathcal{M}} \mathbf{w}\left(\mathbf{p}_{m}\right) \cdot \mathbf{x}\left(\mathbf{p}+\mathbf{p}_{m}+\Delta \mathbf{p}_{m}\right)
\end{equation}

Specifically, we adopt three deformable convolution layers with the kernel size $3\times3$ followed by ReLU activation after each layer. And then a $1\times1$ convolution is leveraged to generate the density map. The final counting number of objects can be computed by summing all the pixel values of the density map. With the dynamic sampling scheme in deformable convolution, the orientation arbitrariness due to the overhead perspective in the remote sensing imagery can be well addressed.

\subsection{Ground truth generation}
We generate the ground truth density maps following the procedure of density map generation in previous works~\cite{zhang2016single,li2018csrnet,chen2019scale}. Assuming that there is an object instance at pixel ${x_i}$, it can be represented by a delta function $\delta ( {x - {x_i}} )$. Therefore, for an image with $N$ annotations, it can be represented as follows:
\begin{equation}
H(x) = \sum\limits_{i = 1}^N {\delta \left( {x - {x_i}} \right)}
\end{equation}

\begin{figure}[!tb]
	\centering
	\centerline{\includegraphics[width=0.45\textwidth]{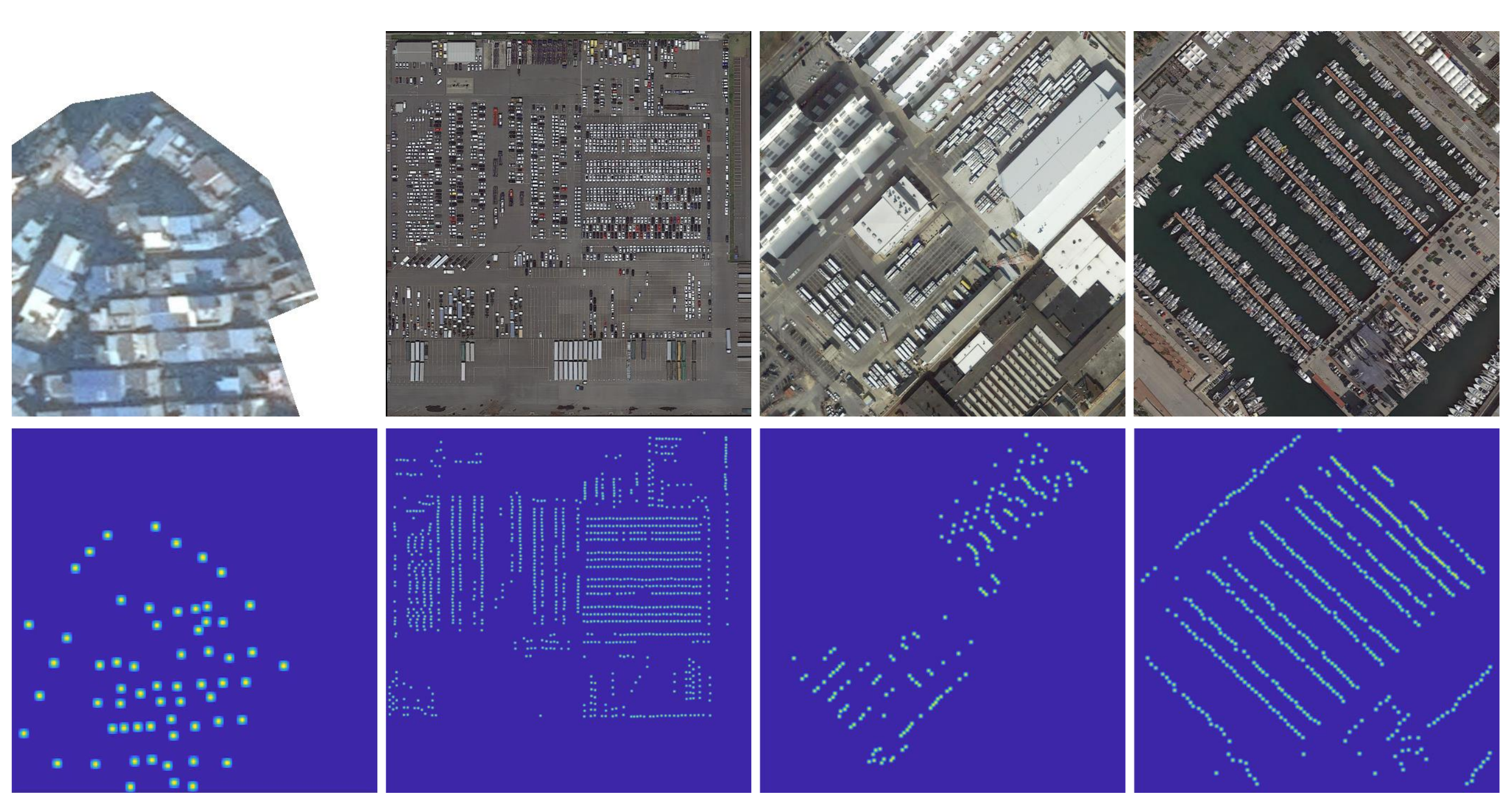}}
	\caption{Visualization of ground truth density maps via Gaussian convolution operation.}
	\label{figure:gauss}
\end{figure}

%Similar as ~\cite{zhang2016single}, geometry-adaptive kernel is also used in our paper,
To generate the density map $F$, we convolute $H(x)$ with a Gaussian kernel, which can be defined as follows:

\begin{equation}
F(x) = \sum\limits_{i = 1}^N {\delta \left( {x - {x_i}} \right)} *{G_{{\sigma _i}}}(x),{\kern 1pt} {\kern 1pt} {\kern 1pt} {\kern 1pt} {\kern 1pt} {\kern 1pt}
\end{equation}
where $\sigma_{i}$ represents the standard deviation. Empirically, we adopt the fixed kernel with $\sigma=15$ for all the experiments. The ground truth density maps after Gaussian convolution operation are visualized in Fig.~\ref{figure:gauss}.

\subsection{Loss function}

We adopt Euclidean distance as loss function to evaluate the difference between the predicted density maps and the ground truths, which is widely adopted by many other counting works~\cite{zhang2016single,boominathan2016crowdnet,sam2017switching}. The $L_2$ loss function is defined as follows:
\begin{equation}
L\left( \Theta  \right) = \frac{1}{{2N}}\sum\limits_{i = 1}^N {\left\| {F\left( {{X_i};\Theta } \right) - F_i^{GT}} \right\|} _2^2
\end{equation}
where $N$ means the batch size, ${X_i}$ represents the input image and $\Theta$ indicates the trainable parameters, $F\left( {{X_i};\Theta } \right)$ and $F_i^{GT}$ are an estimated density map and its corresponding ground truth, respectively.
%%-------------------------------------------------------------------------------------------------------------------------------------------------------------------------------------%%

\section{Remote sensing object counting (RSOC) dataset}
\label{section:dataset}
\begin{figure*}[!tb]
	\centering
	\centerline{\includegraphics[width=1.0\textwidth]{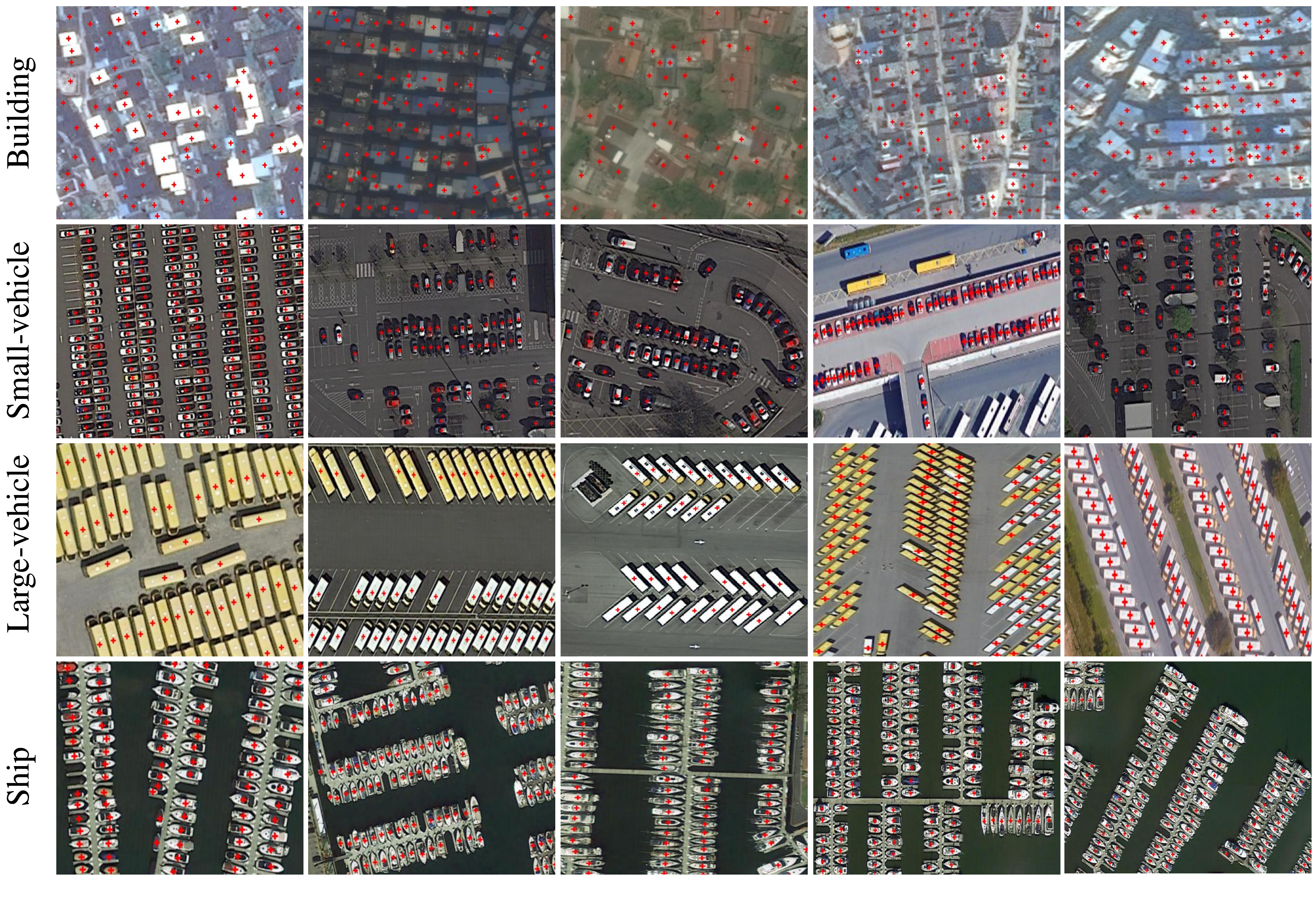}}
	\caption{Example images from the RSOC dataset. Four widely studied and concerned objects including building, small-vehicle, large-vehicle and ship are included in this dataset.}
	\label{fig:images}
\end{figure*}

The absence of publicly available dataset especially large-scale datasets seriously limits the progress of object counting in the remote sensing field. Nowadays, there are only a few datasets available for the community. These datasets are either in a small scale or counting can be achieved easily even using an off-the-shelf detector. For example, the oil trees dataset~\cite{salami2019fly} has only 10 images and 1251 instances in total. The few number of instances easily leads to overfitting for deep learners. In spite of 32,716 cars the COWC dataset~\cite{mundhenk2016large} has, too much contextual information it contains, making it more suitable for detection tasks. CARPK~\cite{hsieh2017drone} is a newly collected drone-based dataset, which has nearly 90,000 cars with bounding box annotations. The distributes of these objects in this dataset are scattered, thus making it more suitable for detection task rather than counting. More recently, a large-scale aerial image based dataset, named DLR-ACD~\cite{bahmanyar2019mrcnet}, is proposed for counting task. This dataset is quite large, which contains 226,291 instances. However, there are only 33 images in it. And more importantly, it is annotated for crowd counting and thus cannot be used for geographic object counting tasks. A simple statistics information of these four datasets is given in Table~\ref{tabel:statistics}. To facilitate counting research in remote sensing community, we construct a large-scale remote sensing object counting dataset, termed RSOC. It consists of 3057 images and 286,539 instances in total. To our best knowledge, this is the largest dataset available for the remote sensing object counting task.  Some representative samples are presented in Fig.~\ref{fig:images}.

%%-------------------------------------------------------------------------------------------------------------------------------------------------------------------------------------%%

\begin{figure*}[!htb]
	\centering
	\centerline{\includegraphics[width=1.0\textwidth]{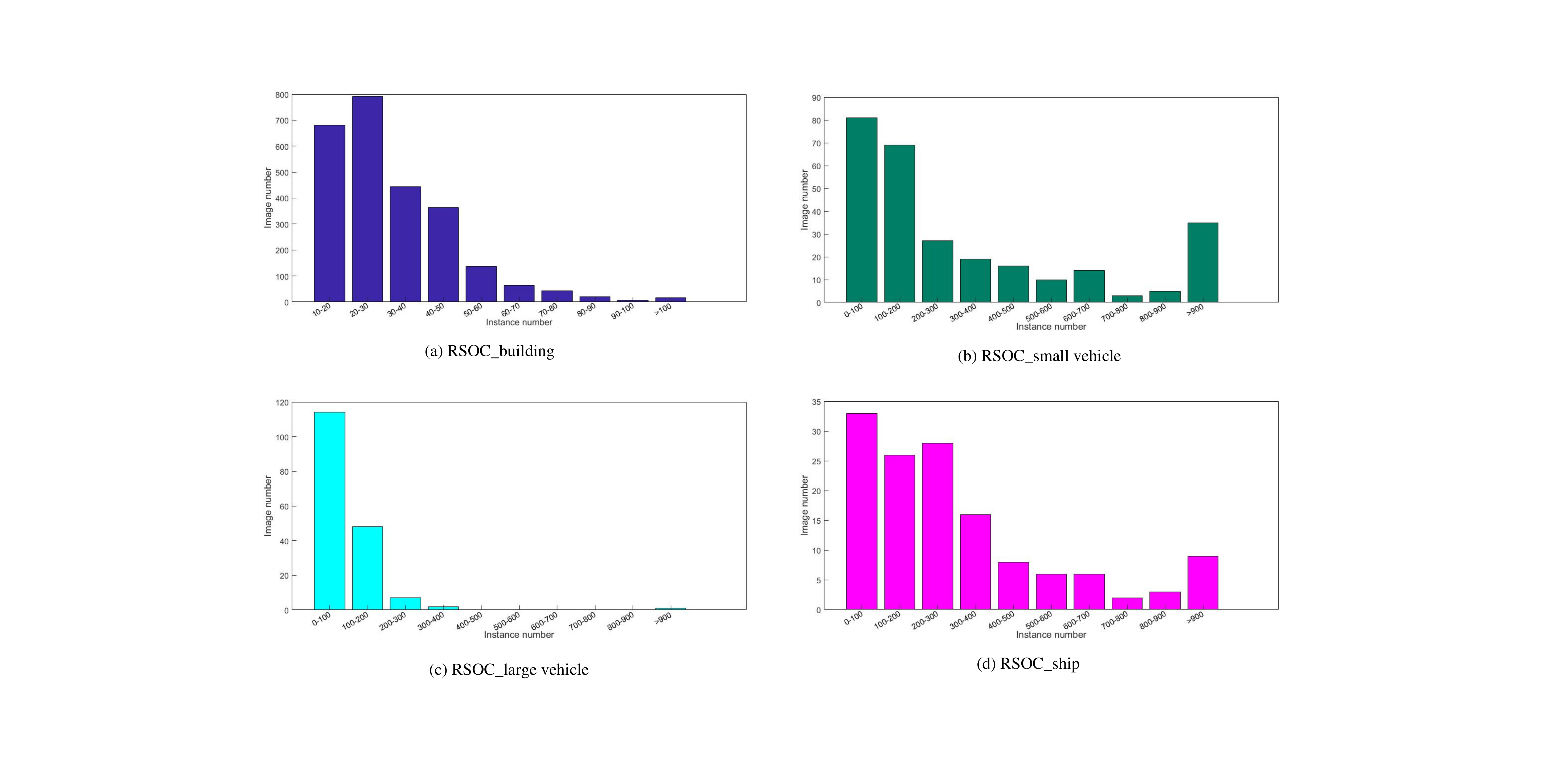}}
	\caption{Distributions of number of instances per image for the RSOC dataset.}
	\label{fig:hist}
\end{figure*}

\textbf{Data collection.} Four types of objects involving buildings, small vehicles, large vehicles, and ships are included in our RSOC dataset, these four types are among the main concerns of researches in the remote sensing community. The images of buildings are collected from Google Earth, while the other three categories are collected from the DOTA dataset~\cite{xia2018dota}, which is a very large dataset built for object detection in aerial images. During collection, the easy cases, i.e., images containing only disperse distributed objects are removed from the RSOC, since we only focus on cluster instances and the count of those disperse ones can be easily inferred from an off-the-shelf detector. As a consequence, there leaves only hard samples such as crowded inshore ships and intensively packed vehicles in the parking lots. Finally, there are 280 images for small vehicles, 172 images for large vehicles, 137 images for ships are included in our dataset. Incorporating 2468 images collected from Google Earth for buildings, the ROSC dataset has 3057 images in total. For each subset, we divide it into training and testing sets as illustrated in Table~\ref{tabel:statistics}. The instance distribution of each subset is plotted in Fig.~\ref{fig:hist}.

\textbf{Annotation.} To reduce the workload and speed up the annotation, the buildings are annotated with center points. Images from DOTA~\cite{xia2018dota} are labeled with rotated quadrilateral bounding boxes enclosing the objects. The labels are denoted as $\left\{\left(x_{i}, y_{i}\right), i=1,2,3,4\right\}$, where $\left(x_{i}, y_{i}\right)$ indicates the positions of the vertices of the boxes in the image. We take the centroid of the bounding box as the central location, which can be calculated as follows,
\begin{equation}
(x, y)=\left(\frac{1}{4} \sum_{i=1}^{4} x_{i}, \frac{1}{4} \sum_{i=1}^{4} y_{i}\right)
\end{equation}

\begin{figure*}[htbp]
\centering

\subfigure[]{
\begin{minipage}[t]{0.75\linewidth}
\centering
\includegraphics[width=5in]{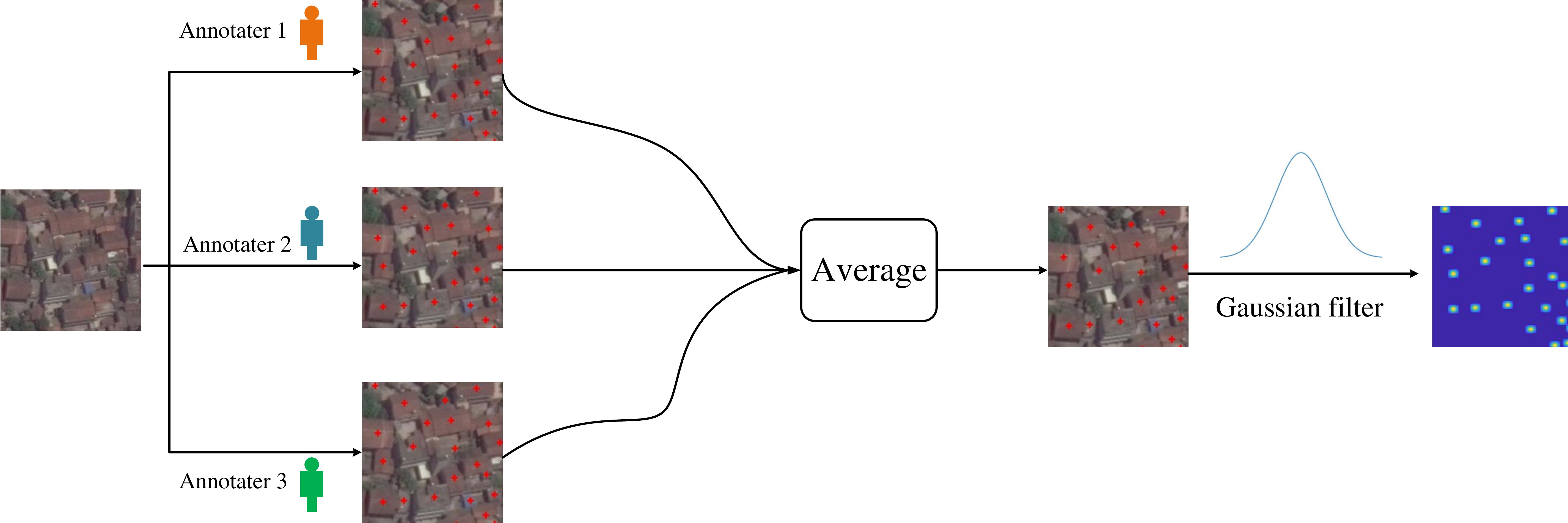}
%\caption{fig1}
\end{minipage}%
}%

                 %这个回车键很重要 \quad也可以
\subfigure[]{
\begin{minipage}[t]{0.75\linewidth}
\centering
\includegraphics[width=5in]{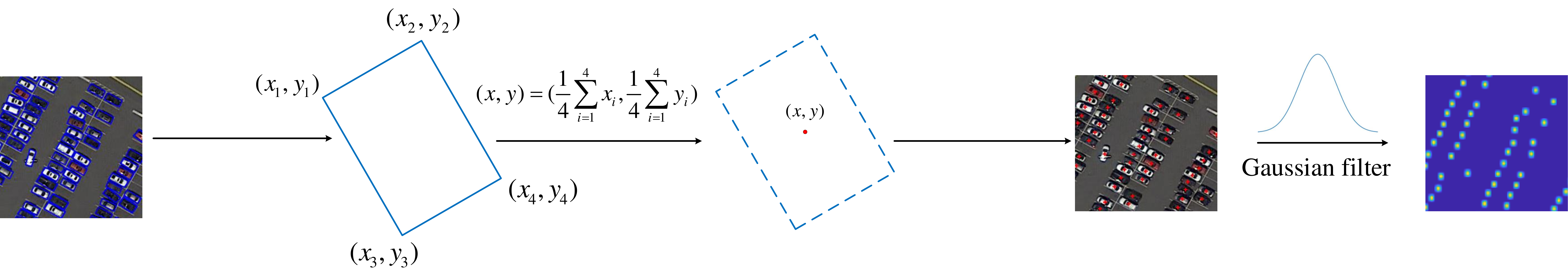}
%\caption{fig2}
\end{minipage}
}%

\centering
\caption{The annotation process for the dataset construction. (a) The annotation with center points for build-ings and (b) The annotation with orientation bounding box, then transform them into center points, the image is derived from DOTA dataset. }
\label{fig:annota}
\end{figure*}

The specific annotation process of our constructed dataset is depicted in Fig.~\ref{fig:annota}.

%%-------------------------------------------------------------------------------------------------------------------------------------------------------------------------------------%%

\textbf{Statistics.} More information on this dataset is given in Table~\ref{tabel:statistics}, from which we can observe that the RSOC dataset has distinct features from other datasets: 1) large data capacity: as mentioned before, the RSOC dataset consists of four categories, 3057 images, and 286,539 instances. It is the largest counting dataset for remote sensing image understanding up to today; 2) large scale variation: the size of objects in the dataset ranges from a dozens of pixels to thousands of pixels, making it extremely challenging for counting; 3) diverse scenes and categories: the dataset covers a variety of scenes including parking lots, towns, villages, harbors and so on. Each scene contains specific annotations such as buildings, vehicles, ships.

%%-------------------------------------------------------------------------------------------------------------------------------------------------------------------------------------%%

\begin{table*}[!htb]
	%\tiny
	\caption{Brief statistical information of the proposed RSOC and other four  counting datasets in the remote sensing filed.}
	%\vspace{-0.3cm}
	\begin{center}
		\renewcommand{\arraystretch}{1.0}	
		\setlength\tabcolsep{5pt}
		\begin{tabular}{c|c|c|c|c|c|c|c|c|c|c}
			
			\hline
			\multicolumn{2}{l}{\multirow {2}{*}{Dataset}} &\multicolumn{1}{|c}{\multirow {2}{*}{Platform}} &\multicolumn{1}{|c}{\multirow {2}{*}{Images}} &\multicolumn{1}{|c}{\multirow {2}{*}{Training/test}} &\multicolumn{1}{|c}{\multirow{2}{*}{Average Resolution}} &\multicolumn{1}{|c}{\multirow {2}{*}{Annotation Format}}
			&\multicolumn{4}{|c}{Count Statistics} \\
			%\hline
			\cline{8-11}
			\multicolumn{2}{c}{}  &\multicolumn{1}{|c}{} &\multicolumn{1}{|c}{} &\multicolumn{1}{|c}{} &\multicolumn{1}{|c}{} &\multicolumn{1}{|c}{} &\multicolumn{1}{|c}{Total}  & Min  &Average &Max \\
			\hline
			\multicolumn{2}{l|}{Olive trees~\cite{salami2019fly}} &UAV &10 &-- &4000$\times$3000 &circle &1251 &107 &125.1 &143 \\
			%\hline
			\multicolumn{2}{l|}{COWC~\cite{mundhenk2016large}} &aerial &-- &-- &low &center point &32,716  &-- &-- &-- \\
			%\hline
			\multicolumn{2}{l|}{CARPK~\cite{hsieh2017drone}} &drone &1448 &989/459 &1280$\times$720 &bounding box &89,777  &1 &62 &188 \\
			%\hline
			\multicolumn{2}{l|}{DLR-ACD~\cite{bahmanyar2019mrcnet}} &aerial &33 &19/14 &3619$\times$5226 &center point &226,291 &285 &6857 &24,368 \\
			\hline
			\hline
            \multirow {4}{*}{RSOC}
			&\multicolumn{1}{|c|}{Building} &satellite &2468 &1205/1263 &512$\times$512 &center point &76,215 &15 &30.88 &142 \\
			%\hline
			&\multicolumn{1}{|c|}{Small-vehicle} &satellite &280 &222/58 &2473 $\times$ 2339 &oriented bounding box &148,838 &17 &531.56 &8531 \\
			%\hline
			&\multicolumn{1}{|c|}{Large-vehicle} &satellite &172 &108/64 &1552 $\times$ 1573 &oriented bounding box &16,594 &12 &96.48 &1336 \\
			%\hline
			&\multicolumn{1}{|c|}{Ship} &satellite &137 &97/40 &2558 $\times$ 2668 &oriented bounding box &44,892 &50 &327.68 &1661 \\
			\hline
		\end{tabular}
	\end{center}
	\label{tabel:statistics}
\end{table*}

\section{Experiments}
\label{section:experimets}
In this section, we benchmark the RSOC dataset by conducting extensive experiments on it. In addition, the ASPD-Net is evaluated on the RSOC and compared with previous state-of-the-art counting methods to demonstrate its superiority. Beyond comparison, ablation studies are also provided to verify the effectiveness of ASPD-Net. Besides, we conduct experiments on one popular crowd counting dataset, i.e., ShanghaiTech dataset~\cite{zhang2016single}, to further demonstrate the robustness and generalization ability of our proposed approach.

\subsection{Evaluation metrics}
Two widely used metrics, Mean Absolute Error (MAE) and Root Mean Squared Error (RMSE), are employed to evaluate the performance of the proposed and comparison methods. MAE measures the accuracy of the model, while RMSE measures the robustness. These two metrics are defined as follows:

\begin{equation}
MAE = \frac{1}{n}\sum\limits_{i = 1}^n {\left| {\hat{C}_i - C_i} \right|}
\end{equation}

\begin{equation}
RMSE = \sqrt {\frac{1}{n}\sum\limits_{i = 1}^n {{{\left| {\hat{C}_i - C_i} \right|}^2}} }
\end{equation}
where $n$ is the number of test images, ${\hat{C}_i}$ denotes the predicted count for the $i$th image and ${C_i}$ indicates the ground truth count.

\begin{figure*}[!htb]
	\centering
	\centerline{\includegraphics[width=1.0\textwidth]{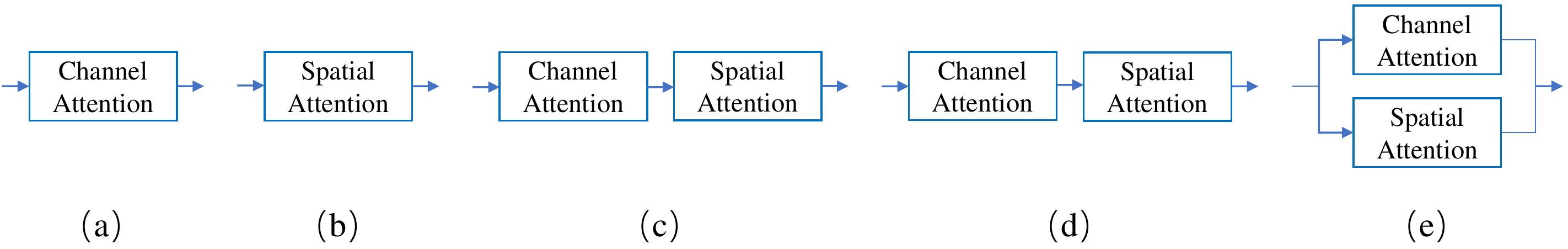}}
	\caption{Configurations of channel- and spatial-attention modules.}
	\label{figure:attarr}
\end{figure*}

\subsection{Implementation details}
We implement the proposed ASPD-Net in PyTorch~\cite{paszke2017pytorch}, train and test it on an NVIDIA 2080Ti GPU. The first 10 convolution layers are fine-tuned from VGG-16~\cite{simonyan2014very}, and the other layers are initialized through a Gaussian noise with 0.01 standard deviation. ASPD-Net is trained in an end-to-end manner. During training, we employ stochastic gradient descent (SGD) to optimize the network and set the learning rate as le-5. For the building subset, we adopt a batch size of 32 and set a batch size of 1 for the other three subsets. For ShanghaiTech crowd counting dataset, we inherit the training manner in \cite{li2018csrnet}. All training will reach convergence in 400 epochs.

We apply data augmentation to generate more training samples. For each image from the training set, 9 sub-images with 1/4 size of the original image are cropped, of which four are adjacent non-overlapping sub-images and the other five are cropped randomly. After then, a mirror flip is applied to double them. Since the ship, large-vehicle and small-vehicle images are with large sizes, which will lead to out-of-memory of GPU in the training phase, therefore, we resize all the large images into 1024 $\times$ 768 pixels before data augmentation.

\subsection{Arrangement of the channel- and spatial-attention}
\label{subsection:arrangement}
\begin{figure}[!htb]
	\centering
	\centerline{\includegraphics[width=0.5\textwidth]{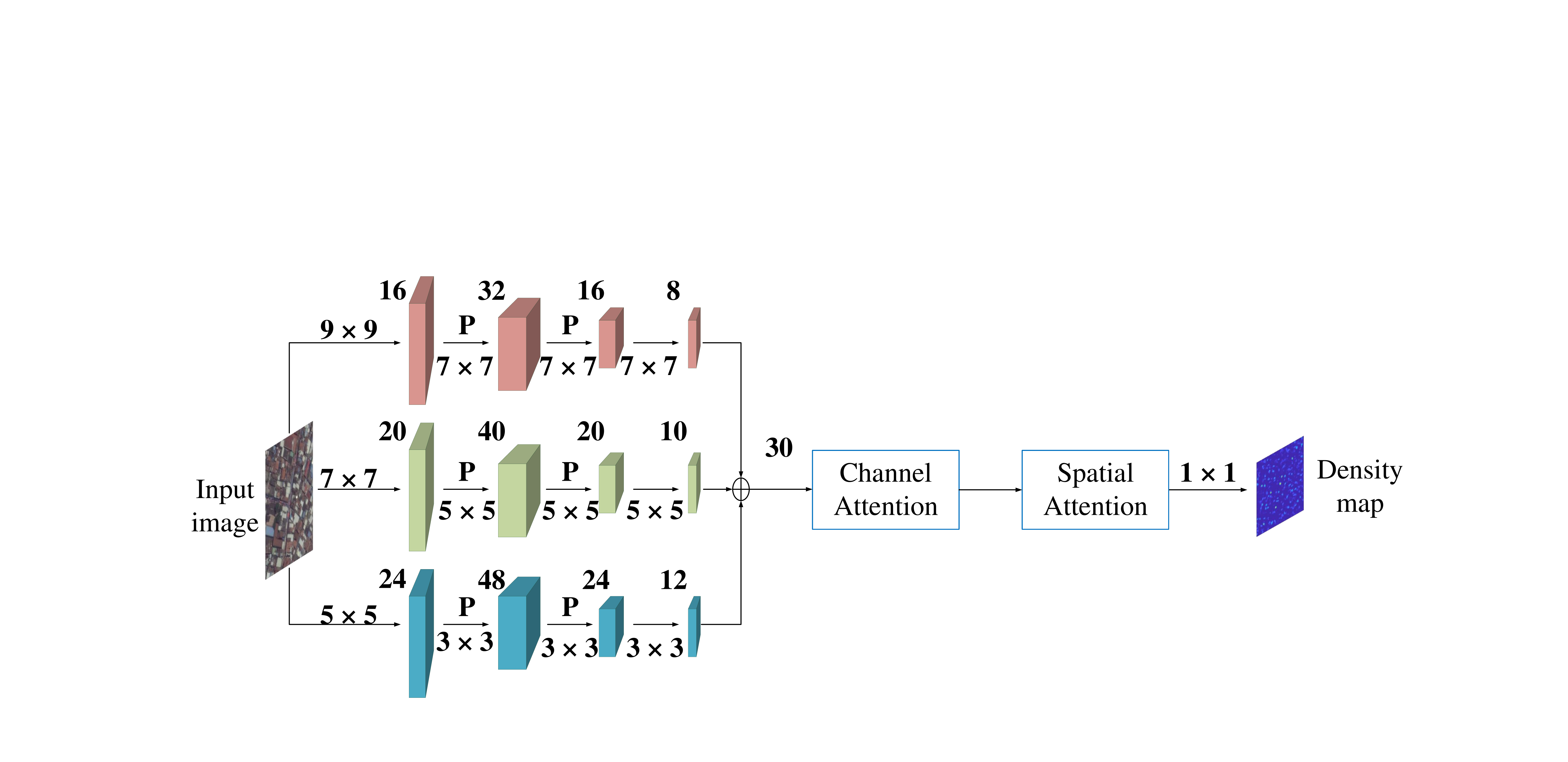}}
	\caption{An example model for evaluating configurations of attention modules.}
	\label{figure:example}
\end{figure}

There are several configurations when integrating channel- and spatial-attention, as shown in Fig.~\ref{figure:attarr}. Here we test all the combinations on top of the MCNN~\cite{zhang2016single} by embedding attention modules into the MCNN. We choose MCNN because it is easy to implement and train. An example test model is shown in Fig.~\ref{figure:example}. We conduct experiments on the RSOC\_building subset. The results are illustrated in Table~\ref{tabel:cbam}. We can find that incorporating attention module, either spatial or channel attention, could significantly improve the performance by a large margin. Channel attention performs slightly better than spatial attention. Combining two attentions together could further boost the performance. And sequential assemblies are superior to parallel one. This is consistent with \cite{woo2018cbam}. The \mbox{`channel+spatial'} configuration obtains the best result, thus in the following experiments, we employ it as the  attention module to embed into the proposed ASPD-Net.

\begin{table}[!h]
	\centering
	\fontsize{8.0}{8}\selectfont
	\begin{threeparttable}
		\caption{Impacts of different attention configurations.}
		\label{tabel:cbam}
		\begin{tabular}{lcc}
			\toprule
			\multirow{2}{*}{Method}&
			\multicolumn{2}{c}{RSOC\_building}\cr
			\cmidrule(lr){2-3}
			&MAE &RMSE \cr
			\midrule
			MCNN~\cite{zhang2016single}&13.65 &18.56  \cr
			MCNN~\cite{zhang2016single}+spatial&11.41 &16.09  \cr
			MCNN~\cite{zhang2016single}+channel&11.20 &15.92  \cr
            MCNN~\cite{zhang2016single}+channel$\&$spatial in parallel&11.21 &15.83  \cr
			MCNN~\cite{zhang2016single}+spatial+channel&11.11 &15.77  \cr
			MCNN~\cite{zhang2016single}+channel+spatial&\bf{10.18} &\bf{14.52} \cr
			\bottomrule
		\end{tabular}
	\end{threeparttable}
\end{table}

\begin{table*}
  \centering
  \fontsize{9.0}{9}\selectfont
  \begin{threeparttable}
  \caption{Ablation experiments on RSOC\_building dataset.}
  \label{table:ablation}
    \begin{tabular}{lcccccccc}
    \toprule
    \multirow{2}{*}{Method}&
    \multicolumn{2}{c}{RSOC\_building} &\multicolumn{2}{c}{RSOC\_small-vehicle} &\multicolumn{2}{c}{RSOC\_large-vehicle} &\multicolumn{2}{c}{RSOC\_ship}\cr
    \cmidrule(lr){2-9}
    &MAE &RMSE &MAE &RMSE &MAE &RMSE &MAE &RMSE\cr
    \midrule
    Baseline&8.00 &11.78 &443.72 &1252.22 &34.10 &46.42 &240.01 &394.81 \cr
    Baseline+Att&7.92 &11.67 &439.51 &1248.95 &28.75 &41.63 &228.45 &365.37\cr
    Baseline+Att+SPM&7.85 &11.58 &436.84 &1243.73 &24.38 &36.54 &211.58 &334.53\cr
    Baseline+Att+SPM+DCM (\textbf{ours})&\textbf{7.59} &\textbf{10.66} &\textbf{433.23} &\textbf{1238.61} &\textbf{18.76} &\textbf{31.06} &\textbf{193.83} &\textbf{318.95} \cr
    \bottomrule
    \end{tabular}
    \end{threeparttable}
\end{table*}

\begin{figure*}[!htb]
	\centering
	\centerline{\includegraphics[width=1.0\textwidth]{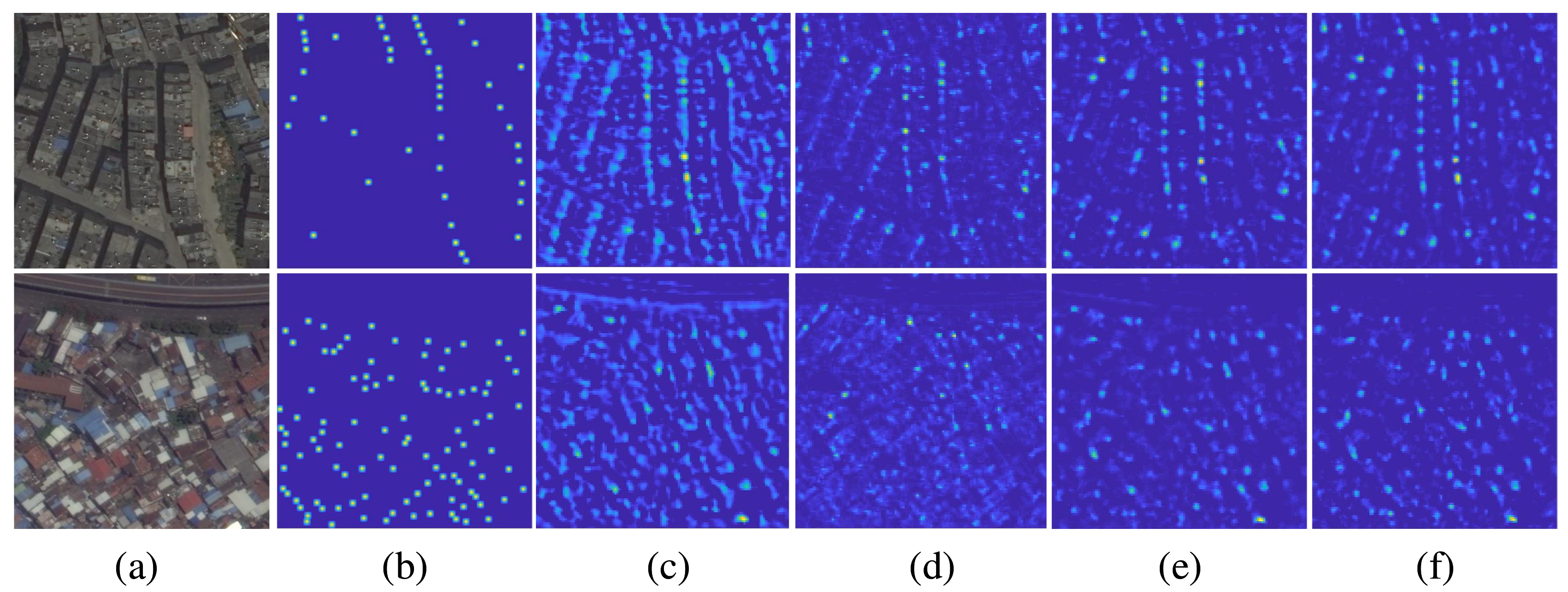}}
	\caption{Density maps generated from each stage of our proposed method. (a) Inputs (b)-(f) represent GT, density maps generated from the methods of Baseline, Baseline+attention, Baseline+attention+SPM and Ours, respectively.}
	\label{figure:stage}
\end{figure*}

\subsection{Ablation study}
To better quantify the contribution of different modules of our method, we conduct an ablation study on the RSOC\_building subset with simplified models.

\noindent $\bullet$ \textbf{Baseline}: We set a baseline method similar to CSRNet~\cite{li2018csrnet}, which is composed of 10 convolution layers carved from VGG-16 and 6 dilated convolution layers with dilation rate 2.

\noindent $\bullet$ \textbf{Baseline+Att}: A sequential channel-spatial attention module (Att) is added on top of the baseline method.

\noindent $\bullet$ \textbf{Baseline+Att+SPM}: The Att module and the scale pyramid module (SPM) are added on the top of the baseline method.

\noindent $\bullet$ \textbf{Baseline+Att+SPM+DCM}: the proposed ASPD-Net.

The results of the ablation experiments are tabulated in Table~\ref{table:ablation}. From the table, we can see that each component in our network contributes to performance improvement. Specifically, the naive baseline method does not perform the most optimal performance. Using the attention module captures the global and local dependencies of the feature maps. The use of scale pyramid module further improves the performance by capturing the multi-scale information. A toy example for the visualization of density maps on RSOC\_building subset is depicted in Fig.~\ref{figure:stage}. It can be intuitively observed that by incorporating attention module, scale pyramid module, and deformable convolution module into the unified framework, our proposed ASPDNet can obtain the superior counting performance and accurate predicts.

\subsection{Comparison with state-of-the-arts on RSOC dataset}
We make comparison with several state-of-the-art counting methods, which are first raised for crowed counting, however, they can be applicable for object counting in remote sensing images. We report the results in Table~\ref{table:comparison}. It can be observed that our approach achieves the best performance on the RSOC dataset. Specifically, compared with the second best state-of-the-art method, ASPDNet improves the performance with 1.94\% MAE and 7.14\% RMSE on RSOC\_building subset, 0.47\% MAE and 0.77\% RMSE on RSOC\_small vehicle subset, 35.40\% MAE and 33.09\% RMSE on RSOC\_large vehicle subset and 3.86\% MAE and 4.18\% RMSE on RSOC\_ship subset, respectively. From the evaluation metrics, we can find that even though we have achieved the best performance on the proposed RSOC dataset, there is still a large margin to be improved, especially for small objects such as small-vehicles and ships. This is consistent with other counting tasks that higher congested the scene is, more challenging the counting task will be. This also indicates the challenging nature of the proposed RSOC is and we hope this will encourage more research efforts to put on our dataset.

\begin{figure*}[!tb]
	\centering
	\centerline{\includegraphics[width=1.0\textwidth]{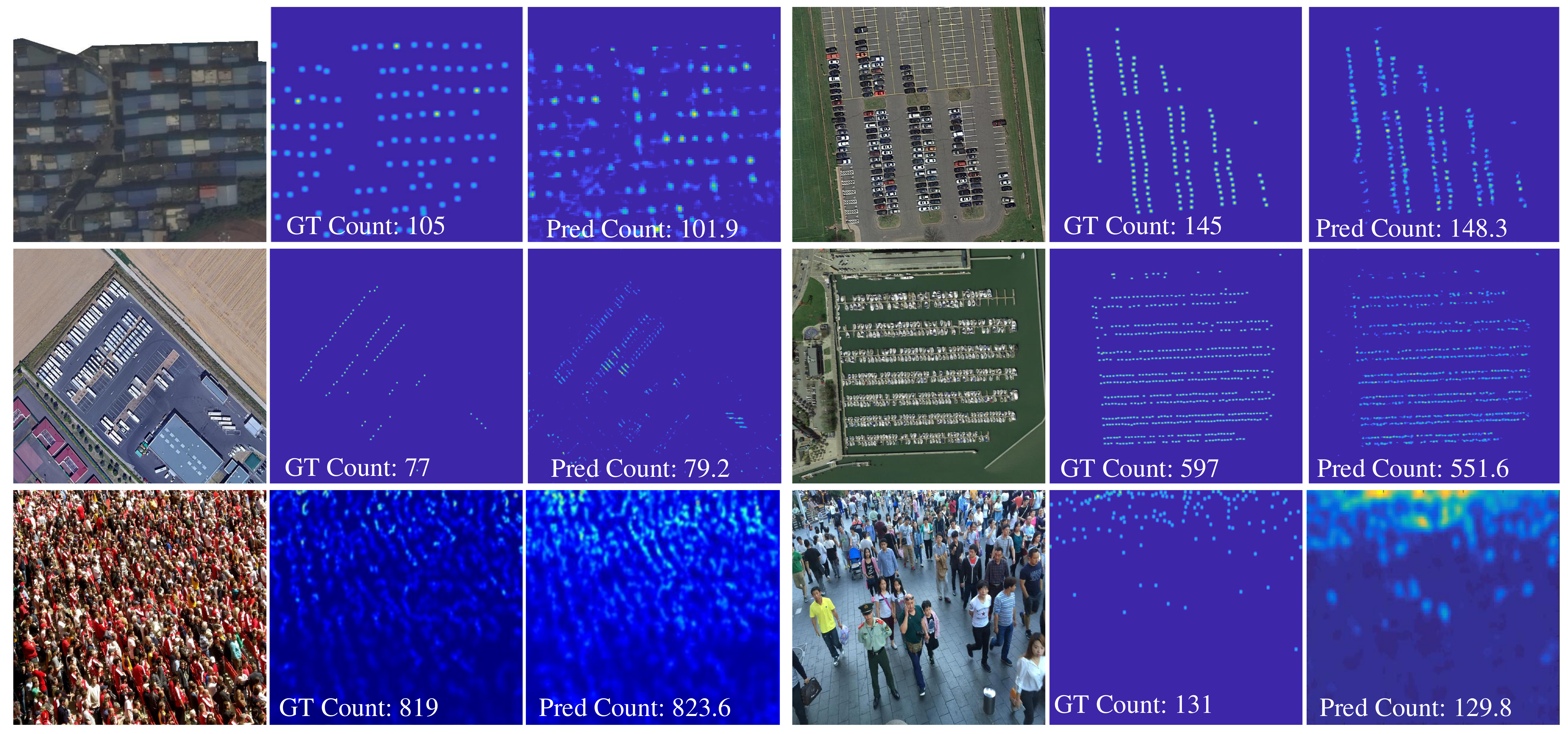}}
	\caption{Visualization of density map for ASPDNet on various datasets. From left to right, for each visualization, the original image, the ground-truth density map, and the predicted density map, respectively. From left to right, top to down, the visualization of RSCO\_building, RSOC\_small-vehicle, RSOC\_large-vehicle, RSOC\_ship, ShanghaiTech Part\_A~\cite{zhang2016single}, and ShanghaiTech Part\_B~\cite{zhang2016single}, respectively.}
	\label{fig:results}
\end{figure*}

Figs.~\ref{fig:results} depicts the visualization results for some sample images from RSOC\_building, RSOC\_small vehicle, RSOC\_large vehicle and RSOC\_ship subsets. It can be observed that our proposed method obtains high-quality density maps with small count errors. It also indicates even under the extreme conditions of scale variation, complex clutter backgrounds, and orientation arbitrariness, our proposed approach still performs strong robustness capacity. Meanwhile, from the generated density maps, we can find that our model has accurate localization ability to some extent.

\begin{table*}[!htb]
  %\tiny
  \caption{Performance comparison on our constructed RSOC dataset. The top three performances are highlighted with \textbf{bold}, \uline{underline} and \uwave{under wave}.}
  \vspace{-0.3cm}
	\begin{center}
		\resizebox{\textwidth}{!}{
		\begin{tabular}{r|rc|cc|cc|cc}
			\hline
			\multirow{2}{*}{\backslashbox{Methods}{Datasets}} & \multicolumn{2}{c|}{Building}& \multicolumn{2}{c|}{Small vehicle}&\multicolumn{2}{c|}{Large vehicle}&\multicolumn{2}{c}{Ship}\\
			\cline{2-9}
			&MAE &RMSE &MAE &RMSE &MAE &RMSE &MAE &RMSE  \\
			\hline
            MCNN~\cite{zhang2016single} &13.65 &16.56 &488.65 &1317.44 &36.56 &55.55 &263.91 &412.30 \\
            CMTL~\cite{sindagi2017cnn} &12.78 &15.99 &490.53 &1321.11 &61.02 &78.25 &251.17 &403.07 \\
            CSRNet~\cite{li2018csrnet} &\uwave{8.00} &11.78 &443.72 &\uwave{1252.22} &34.10 &\uline{46.42} &\uwave{240.01} &394.81 \\
            SANet~\cite{cao2018scale} &29.01 &32.96 &497.22 &1276.66 &62.78 &79.65 &302.37 &436.91 \\
            SFCN~\cite{wang2019learning} &8.94 &12.87 &\uwave{440.70} &\uline{1248.27} &\uwave{33.93} &49.74 &240.16 &394.81 \\
            SPN~\cite{chen2019scale} &\uline{7.74} &\uline{11.48} &445.16 &1252.92 &36.21 &50.65 &241.43 &\uwave{392.88} \\
            SCAR~\cite{gao2019scar} &26.90 &31.35 &497.22 &1276.65 &62.78 &79.64 &302.37 &436.92 \\
            CAN~\cite{liu2019context} &9.12 &13.38 &457.36 &1260.39 &34.56 &49.63 &282.69 &423.44 \\
            SFANet~\cite{zhu2019dual} &8.18 &\uwave{11.75} &\uline{435.29} &1284.15 &\uline{29.04} &\uwave{47.01} &\uline{201.61} &\uline{332.87} \\
            ASPDNet (\textbf{proposed}) &\textbf{7.59} &\textbf{10.66} &\textbf{433.23} &\textbf{1238.61} &\textbf{18.76} &\textbf{31.06} &\textbf{193.83} &\textbf{318.95} \\
        \hline
		\end{tabular}}
	\end{center}
\label{table:comparison}
\end{table*}

\subsection{Comparison with state-of-the-arts on crowd counting dataset}
To further demonstrate the effectiveness and generalization ability of our proposed method, we apply it on one crowd counting dataset, i.e., ShanghaiTech dataset~\cite{zhang2016single}. This dataset is composed of 1,198 annotated images with a total of 330,165 presidents, which is split into two parts, Part\_A and Part\_B. Part\_A contains 482 highly congested images which are randomly crawled from the Internet, where 400 images serve as training and the remaining are for testing. Part\_B contains 716 image with relatively sparse people which are taken from the busy streets of metropolitan areas in Shanghai, China, where 400 images are for training and the rest 316 are for testing.

We report the comparison results in Table~\ref{table:crowd}. It can be seen that compared with the state-of-the-art methods, our proposed method can obtain the best results, which demonstrates great robustness and generalization ability. Specifically, compared with the baseline method, CSRNet~\cite{li2018csrnet}, it gains 10.85\% / 16.35\% (MAE / RMSE) on Part\_A, and 32.08\% / 34.38\% (MAE / RMSE) on Part\_B, respectively. Some visualizations of density maps are depicted in Fig.~\ref{fig:results}, which shows the proposed method still can generate accurate predicts for diverse objects from sparse to dense scenarios.

\begin{table}[!htb]
  %\tiny
  \caption{Performance comparison on ShnaghaiTech crowd counting dataset. The best performance is highlighted with \textbf{bold}.}
  \vspace{-0.3cm}
	\begin{center}
		\renewcommand{\arraystretch}{1.0}	
		\setlength\tabcolsep{3pt}
        {
		\begin{tabular}{|r|cc|cc|cc|cc}
			\hline
			\multirow{2}{*}{\backslashbox{Methods}{Datasets}} &\multicolumn{2}{c|}{ShanghaiTech Part\_A} &\multicolumn{2}{c|}{ShanghaiTech Part\_\_B}\\
			\cline{2-5}
			&MAE &RMSE &MAE &RMSE  \\
			\hline
            MCNN~\cite{zhang2016single} &110.2 &173.2 &26.4 &41.3 \\
            CMTL~\cite{sindagi2017cnn}  &101.3 &152.4 &20.0 &31.1 \\
            Switch-CNN~\cite{sam2017switching} &90.4 &135.0 &21.6 &33.4 \\
            IGCNN~\cite{babu2018divide} &72.5 &118.2 &13.6 &21.1 \\
            CSRNet~\cite{li2018csrnet}  &68.2 &115.0 &10.6 &16.0  \\
            SANet~\cite{cao2018scale}  &67.0 &104.5 &8.4 &13.6 \\
            SFCN~\cite{wang2019learning} &64.8 &107.5 &7.6 &13.0  \\
            SPN~\cite{chen2019scale} &61.7 &99.5 &9.4 &14.4 \\
            SCAR~\cite{gao2019scar}  &66.3 &114.1 &9.5 &15.2 \\
            ADCrowdNet~\cite{liu2019adcrowdnet} &70.9 &115.2 &7.7 &12.9\\
            BL~\cite{ma2019bayesian} &62.8 &101.8 &7.7 &12.7 \\
            %CAN~\cite{liu2019context} &62.3 &100.0 &7.8 &12.2\\
%            SFANet~\cite{zhu2019dual}  &59.8 &99.3 &6.9 &10.9 \\
            ASPDNet (\textbf{proposed}) &\textbf{60.8} &\textbf{96.2} &\textbf{7.2} &\textbf{10.5} \\
        \hline
		\end{tabular}}
	\end{center}
\label{table:crowd}
\end{table}

\subsection{Standard deviations experiments}
To validate the stability of our proposed method, following~\cite{sindagi2020learning}, we also report the standard deviations of our methods on the constructed dataset. See Table~\ref{table:standard} for detail, note that the standard deviations are computed using 5 trials.

\begin{table*}[!htb]
  %\tiny
  \caption{Standard deviations of our methods on the four categories. }
  \vspace{-0.3cm}
	\begin{center}
		\resizebox{\textwidth}{!}{
		\begin{tabular}{r|rc|cc|cc|cc}
			\hline
			\multirow{2}{*}{\backslashbox{Methods}{Datasets}} & \multicolumn{2}{c|}{Building}& \multicolumn{2}{c|}{Small vehicle}&\multicolumn{2}{c|}{Large vehicle}&\multicolumn{2}{c}{Ship}\\
			\cline{2-9}
			&MAE &RMSE &MAE &RMSE &MAE &RMSE &MAE &RMSE  \\
			\hline
            Ours &7.59$\pm$0.8 &10.66$\pm$1.7 &433.23$\pm$0.6 &1238.61$\pm$1.3 &18.76$\pm$0.5 &31.06$\pm$1.2 &193.83$\pm$0.7 &318.95$\pm$1.5 \\
        \hline
		\end{tabular}}
	\end{center}
\label{table:standard}
\end{table*}

\section{Conclusion and future work}
\label{section:conclusion}

Counting the object instance in remote sensing images is a remarkable significant yet scientifically challenging topic. To achieve this, we devise an ASPD-Net by incorporating attention mechanism, scale pyramid module, and deformable convolution module into a unified framework. Moreover, considering that the development of this field has been limited mainly due to the scarcity of large-scale datasets with accurately annotated. To remedy this, we present a new large-scale remote sensing object counting dataset which encompasses 4 object categories with 3057 images, and 286,539 instances in total. Extensive experimental results including quantitative and qualitative demonstrate the effectiveness and superiority of our proposed approach compared with the off-the-shelf state-of-the-art methods for crowd counting. In addition, to further validate the effectiveness of each component and the generalization ability of our designed ASPD-Net, some ablation studies on our constructed RSOC dataset and experiments on one widely used crowd counting dataset are also implemented. We expect that our contribution can bridge the gap and guide the new developments on the object counting in remote sensing imagery.

Nevertheless, there are some drawbacks such as class unbalanced in our proposed RSOC dataset, suboptimal performance on small object counting of the proposed method. Therefore, in the future, we plan to collect more images from various platforms, and devise better model to alleviate the small object counting problems. Meanwhile, we intend to design dedicated class-agnostic algorithms for remote sensing object counting.

\normalem
\bibliographystyle{IEEEtran}
\bibliography{IEEEabrv,references}

\begin{IEEEbiography}[{\includegraphics[width=1.25in,height=1in,clip,keepaspectratio]{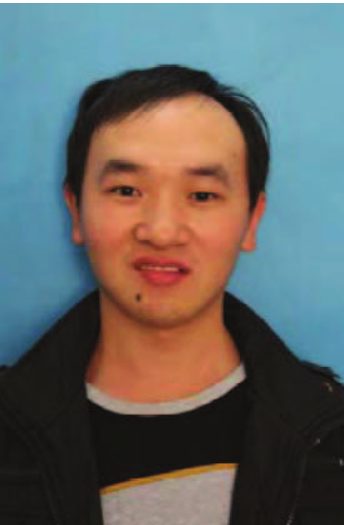}}]{Guangshuai Gao}
received the B.Sc. degree in applied physics from college of science and the M.Sc. degree in signal and information processing from the School of Electronic and Information Engineering, from the Zhongyuan University of Technology, Zhengzhou, China, in 2014 and 2017, respectively.

He is currently pursuing the Ph.D. degree with the Laboratory of Intelligent Recognition and Image Processing, Beijing Key Laboratory of Digital Media, School of Computer Science and Engineering, Beihang University. His research interests include image processing, pattern recognition, remote sensing image analysis and digital machine learning.
\end{IEEEbiography}

\begin{IEEEbiography}[{\includegraphics[width=1.25in,height=1in,clip,keepaspectratio]{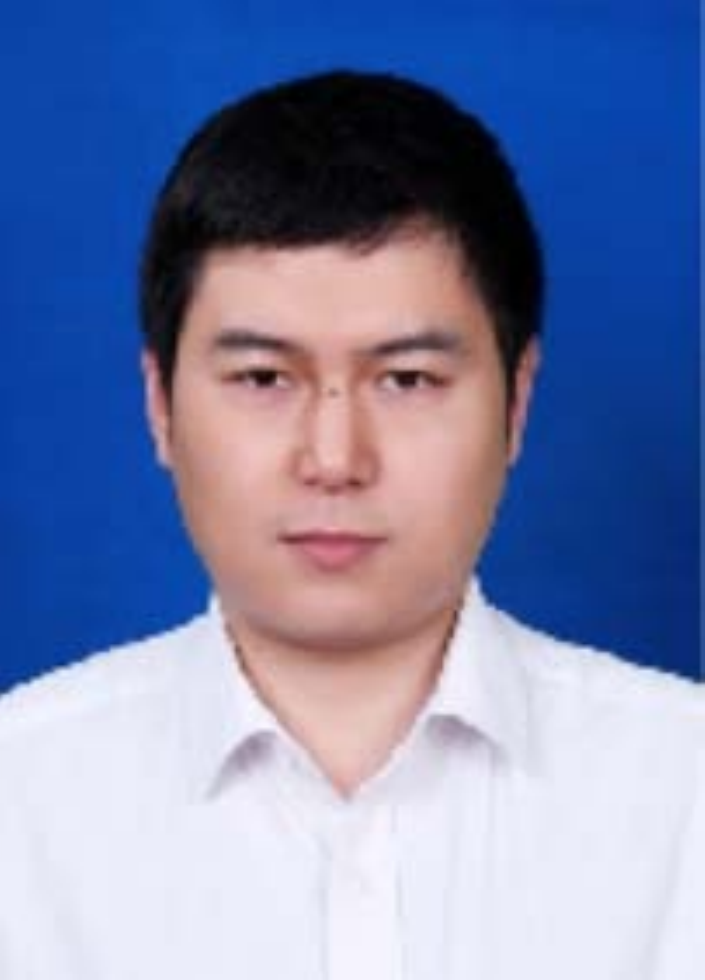}}]{Qingjie Liu}
received the B.S. degree in computer science from Hunan
University, Changsha, China and the Ph.D. degree in computer science from
Beihang University, Beijing, China.

He is currently an Assistant Professor with the School of Computer Science and Engineering, Beihang University.
He is also a Distinguished Research Fellow with the Hangzhou Institute
of Innovation, Beihang University, Hangzhou. His current research interests
include remote sensing image analysis, pattern recognition, and computer
vision. He is a member of the IEEE.
\end{IEEEbiography}

% if you will not have a photo at all:
\begin{IEEEbiography}[{\includegraphics[width=1.25in,height=1in,clip,keepaspectratio]{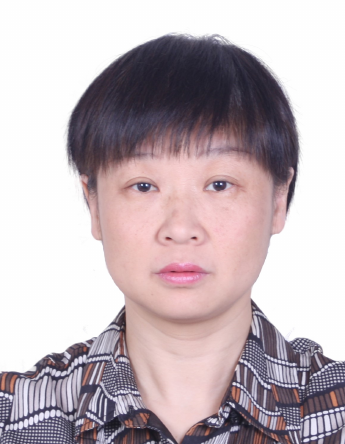}}]{Yunhong Wang}
received the B.S. degree from Northwestern Polytechnical University, Xi’an, China, in 1989, and the M.S. and Ph.D. degrees from the Nanjing University of Science
and Technology, Nanjing, China, in 1995 and 1998, respectively, all in electronics engineering.

She was with the National Laboratory of Pattern Recognition, Institute of Automation, Chinese Academy of Sciences, Beijing, China, from 1998 to 2004. Since 2004, she has been a Professor with the School of Computer Science and Engineering, Beihang University, Beijing, where she is currently the Director of Laboratory of Intelligent Recognition and Image Processing, Beijing Key Laboratory of Digital Media. Her research results have published at prestigious journals and prominent conferences, such as the IEEE TRANSACTIONS ON PATTERN ANALYS IS AND MACHINE INTELLIGENCE (TPAMI), TRANSACTIONS ON IMAGE PROCESSING (TIP), TRANSACTIONS ON INFORMATION FORENSICS AND SECURITY (TIFS), Computer Vision and Pattern Recognition (CVPR), International Conference on Computer Vision (ICCV), and European Conference on Computer Vision (ECCV). Her research interests include biometrics, pattern recognition, computer vision, data fusion, and image processing. She is a Fellow of the IEEE.
\end{IEEEbiography}

\end{document}